\documentclass[letterpaper, 10 pt, conference]{ieeeconf}  % Comment this line out if you need a4paper

\IEEEoverridecommandlockouts                              % This command is only needed if 
\usepackage[utf8]{inputenc} % allow utf-8 input
\usepackage[T1]{fontenc}    % use 8-bit T1 fonts
\usepackage[hidelinks]{hyperref}       % hyperlinks
\usepackage{url}            % simple URL typesetting
\usepackage{booktabs}       % professional-quality tables
\usepackage{amsfonts}       % blackboard math symbols
\usepackage{nicefrac}       % compact symbols for 1/2, etc.
\usepackage{microtype}      % microtypography
\usepackage[separate-uncertainty=true,detect-mode=true]{siunitx} % For si units
\usepackage{mathtools}
\usepackage{amssymb}
\usepackage{amsmath}
\usepackage{tabu}
\usepackage{color}
\usepackage{algorithm}
\usepackage{algorithmic}
\usepackage{hyperref}
\usepackage[capitalize]{cleveref}
\usepackage[pdftex]{graphicx}

\usepackage{svg}
\usepackage{subcaption}
\usepackage[style=ieee,backend=bibtex,url=false,doi=false,natbib=true,mincitenames=1,dashed=false,isbn=false]{biblatex}
 \AtEveryBibitem{\clearfield{issn}}
 \AtEveryCitekey{\clearfield{issn}}
 \usepackage[keeplastbox]{flushend}
 \usepackage{anyfontsize}
\definecolor{colorBluish}{RGB}{101,161,216}
\definecolor{colorBluePastel}{RGB}{139,185,226}
\definecolor{colorGreenish}{RGB}{106,130,94}
\definecolor{colorGreenPastel}{RGB}{150,194,154}
\definecolor{colorRedish}{RGB}{140,21,21}
\definecolor{colorRedPastel}{RGB}{164,117,129}
\definecolor{colorOrangish}{RGB}{237,125,49}
\definecolor{colorOrangePastel}{RGB}{213,169,143}
\definecolor{colorPurplePastel}{RGB}{216,191,216}
\definecolor{colorPurplePastel}{RGB}{216,191,216}

\definecolor{vandeusen}{RGB}{73,92,111}
\definecolor{cordovan}{RGB}{152,68,71}

\definecolor{colorA}{HTML}{AAAAAA}
\definecolor{colorB}{HTML}{999999}
\definecolor{colorC}{HTML}{777777}
\definecolor{colorD}{HTML}{555555}
\definecolor{colorE}{HTML}{444444}
\definecolor{colorF}{HTML}{222222}
\definecolor{colorG}{HTML}{000000}
\definecolor{colorH}{HTML}{999999}
\definecolor{colorI}{HTML}{777777}

\newcommand{\todo}[1]{\textbf{[[#1]]}}

\usepackage{array}
\usepackage{graphicx}
\usepackage{multirow}

\newcommand\MyBox[2]{
  \fbox{\lower0.75cm
    \vbox to 1.7cm{\vfil
      \hbox to 1.7cm{\hfil\parbox{1.4cm}{#1\\#2}\hfil}
      \vfil}%
  }%
}

\newcommand{\liam}[1]{\textcolor{black}{#1}}
\newcommand{\raunak}[1]{\textcolor{black}{#1}}
\newcommand{\paren}[1]{\mathopen{}\mathclose\bgroup\left(#1\aftergroup\egroup\right)}
\newcommand{\brock}[1]{\mathopen{}\mathclose\bgroup\left[#1\aftergroup\egroup\right]}
\newcommand{\curly}[1]{\mathopen{}\mathclose\bgroup\left\{#1\aftergroup\egroup\right\}}

\usepackage{amsmath,amssymb,mathtools}

\newcommand{\suchthat}{\;\ifnum\currentgrouptype=16 \middle\fi|\;}

\DeclarePairedDelimiterX{\infdivx}[2]{(}{)}{%
  #1\;\delimsize\|\;#2%
}

\input{macros.tex}

\title{\LARGE \bf
A Hybrid Rule-Based and Data-Driven Approach to \\
Driver Modeling through Particle Filtering 
}

\author{Raunak Bhattacharyya, Soyeon Jung, Liam Kruse, Ransalu Senanayake, and Mykel J. Kochenderfer% <-this % stops a space
\thanks{R. Bhattacharyya, S. Jung, L. Kruse, R. Senanayake, and M.J. Kochenderfer are with the Stanford Intelligent Systems Laboratory in the Department of Aeronautics and Astronautics at Stanford University, Stanford, CA 94305, USA (email: \{raunakbh, soyeonj, lkruse, ransalu, mykel\}@stanford.edu\}). }
}

\begin{document}
\maketitle
\thispagestyle{empty}
\pagestyle{empty}

\begin{abstract}

% Autonomous vehicles need extensive safety validation in simulation before they can be deployed in the real world.
% To generate sufficient confidence in safety validation using simulation, we need reliable models of human driving behavior to generate realistic scenarios to test autonomous vehicles.
\raunak{Autonomous vehicles need to model the behavior of surrounding human driven vehicles to be safe and efficient traffic participants.}
Existing approaches to modeling human driving behavior have relied on both data-driven and rule-based methods. 
While data-driven models are more expressive, rule-based models are interpretable, which is an important requirement for safety-critical domains like driving.
However, rule-based models are not sufficiently representative of data, and data-driven models are yet unable to generate realistic traffic simulation due to unrealistic driving behavior such as collisions.
In this paper, we propose a methodology that combines rule-based modeling with data-driven learning. 
While the rules are governed by interpretable parameters of the driver model, these parameters are learned online from driving demonstration data using particle filtering.
We perform driver modeling experiments on the task of highway driving and merging using data from three real-world driving demonstration datasets.
Our results show that driver models based on our hybrid rule-based and data-driven approach can accurately capture real-world driving behavior.
Further, we assess the realism of the driving behavior generated by our model by having humans perform a ``driving Turing test,'' where they are asked to distinguish between videos of real driving and those generated using our driver models.

% Modeling human behavior is important for autonomous agents to work with and around humans. These models are required both for the planning process of autonomous agents and for simulations that can replicate the intended environment of their operation. A variety of black-box models of human behavior have been studied recently, but they lack interpretability, which can be important in safety-critical domains such as autonomous driving. Rule-based models can be better suited for providing interpretable models of human behavior, but their heuristics can often be brittle. In this work, we propose a methodology that combines rule-based modeling with data-driven learning. The parameters of an underlying rule-based model are learned online from demonstration data using particle filtering. We also incorporate the inherent stochasticity of human behavior as part of the model parameters. Our method results in a distribution over model parameters that can be sampled from to generate novel scenarios. We demonstrate our method on human driver modeling, a case study inspired from the autonomous driving literature. Our results show that our method is able to accurately capture driving behavior and also generate statistically representative driving scenarios. We assess the realism of these scenarios by having humans perform a ‘driving Turing test’, where they are asked to distinguish between videos of real driving and those generated using our driver models.
\end{abstract}
\section{Introduction}
\raunak{
Driver models are needed for designing safe and efficient autonomous driving systems.
        Autonomous vehicles can use these models to make predictions about the behavior of surrounding human drivers.
            In addition, these models can be used to support the validation of autonomous driving systems.
                Evaluating autonomous vehicles on real-world drive tests is time-consuming, expensive, and potentially dangerous. 
                        Validation through simulation provides a promising alternative to real-world testing, but simulations must be based on realistic models of human drivers.
}

Driver modeling is characterized by a high degree of uncertainty.
The behavior of any given \raunak{driver} depends on a multitude of unobservable psychological and physiological factors, e.g., the driver's latent objectives and unique ``driving style.''
    Modeling is further complicated by interaction between multiple drivers.
        Even if all other sources of uncertainty in a traffic scene are ignored, this interaction between decision-making agents yields a complex multi-modal distribution over possible outcomes that can be very challenging to model.

Existing approaches to modeling human behavior have relied on both black-box and rule-based methods. 
\emph{Black-box} models arising out of purely data-driven methods (e.g., Gaussian mixture models and neural networks \cite{lefevre2014comparison}, \cite{morton2016analysis}, \cite{kuefler2017imitating}, \cite{bhattacharyya2019simulating}) often have the expressive power to capture nuanced driving behavior.
    However, such models lack interpretability and often exhibit unrealistic, even dangerous behavior (e.g., colliding with other vehicles) in regions of the state-space that are underrepresented in the training dataset.
Though usually less expressive than black-box models, \emph{rule-based} models (e.g., the Intelligent Driver Model \cite{treiber2000congested}) are interpretable and---in many cases---can guarantee ``good behavior'' (e.g., collision-free driving).
    This ``good behavior'' arises directly from the model structure itself, which is informed by expert knowledge and applies even in regions of the state space that are underrepresented in the data.
        However, rule-based models are generally determinstic and do not take advantage of the variability we see in large datasets, instead relying on heuristics to assign the parameters of the model~\cite{helbing1995social, treiber2000congested}.

For rule-based models, the model parameters can be selected \emph{offline} or \emph{online}.
    Offline methods can make use of arbitrary amounts of data. These methods usually yield ``average'' parameters for the population of drivers represented in the data set since they aggregate data obtained from all the drivers. Offline estimation is the paradigm of choice for essentially all black-box models and many rule-based models.
    In contrast, online methods can capture idiosyncrasies of individual drivers because these methods use real-time sensor information to select and/or update model parameters.
    The time and information implications of near real-time operation mean that online methods are best-suited to models (i.e., rule-based models) with relatively few parameters.

In this work, we propose a methodology that combines rule-based modeling with data-driven learning. The parameters of an underlying rule-based model are learned online from human demonstration data using particle filtering. The proposed methodology is especially suited to human modeling because human behavior is inherently stochastic, i.e., given the same situation, humans may not necessarily take the same action every time \cite{pentland1999modeling,bansal2019beyond}. In our method, we incorporate this stochasticity as part of model parametrization. Further, our method results in a distribution over model parameters that can be sampled from to generate novel scenarios.

Given driving demonstration trajectories from highway driving scenarios, we recover driver models using particle filtering to infer the parameters of an underlying rule-based model with stochasticity. 
Our results show that our method is able to accurately model human driving trajectories. Subsequently, we use the learned driver models to generate synthetic driving scenarios. We assess the realism of these scenarios by having humans perform a ``driving Turing test'' on the generated driving behavior.

Our contributions are as follows:
\begin{itemize}
    \item Given trajectories of demonstrations from multiple humans and a parametrized rule-based model of human behavior, we provide a methodology to estimate a distribution over the parameters of this model using particle filtering. The procedure incorporates a distribution over the stochasticity inherent to human behavior. The appeal of this method is its simplicity and ease of implementation, making it suitable for real-time driving simulation.
    \item We demonstrate our methodology on the problem of naturalistic driver modeling. We model how humans perform two driving tasks: drive on highways, and perform merging. We use demonstrations from three real-world driving datasets: NGSIM, HighD, and Interaction. Our results show that we are able to accurately model driving behavior based on the root mean square error (RMSE) between demonstration and rollout trajectories. Further, we are able to generate novel driving scenarios as assessed by a "driving Turing test" where human participants are asked to distinguish between real and synthetic driving behavior.
\end{itemize}
\section{Background}

Rule-based modeling is an approach that uses a set of rules that indirectly specifies a mathematical model. For instance, if an autonomous vehicle equipped with a rule-based driving model observes an orange traffic light while driving, its rule enforces the model to set its acceleration to zero. 
% Each term or submodule in a rule-based model has a physical meaning and serves a dedicated task. 
Although rule-based models are interpretable, they are brittle and struggle to generalize to diverse scenarios since they only rely on predefined rules. 
% These models are also not capable of learning the inherent stochasticity and the large amount of variability of the environment they operate in. 
For instance, in autonomous driving, different drivers inherently have different driving patterns~\cite{brown2020modeling}. Therefore, we need a model that generalizes human behavior while accounting for individual variations.

In contrast to rule-based models, we can also develop a model to learn purely from data. Such black-box models do not have a prescribed set of rules. The advantage of data-driven models is that they can learn arbitrarily complex patterns from large amounts of data. However, they have two main disadvantages. First, it can be challenging to explicitly incorporate physical knowledge or structure in such models. For instance, as humans, we know that vehicles should not collide with each other. However, as found by \citet{bhattacharyya2020modeling}, it is challenging for completely data-driven techniques to learn such rules. Second, because the data-driven models are not interpretable, it is difficult to verify and validate them, making them less attractive for safety-critical applications such as autonomous driving. In such applications, engineers should be able to stress test the system before deployment by taking into account various possible failure modes \cite{lee2018}. In case of failure, they should be able to understand the underlying reason for the failure. In the field of system identification, attempts to combine black-box models with white-box models are known as gray-box modeling. A common approach is to combine a partial theoretical structure with data to complete the model. Such models have been used for modeling nonlinear system dynamics \cite{pearson2000gray,gupta2020structured,menda2020scalable}.

With the aim of operating robots around humans, both rule-based and data-driven techniques have been used to model human behavior. \cite{wang2013probabilistic} and \cite{volz2016data} attempt to infer the intentions of humans in a collaborative robot manipulation task using Bayesian estimation and pedestrian crossing using convolutional neural networks, respectively. Such signals about how humans would behave in the future can help robot decision-making \cite{wang2013probabilistic,bai2015intention}. In a similar problem, \cite{chandra2020forecasting} attempts to model future actions given the observations in previous time steps using recurrent neural networks. Some of these models are purely rule-based~\cite{helbing1995social} while some are purely data-driven~\cite{chandra2020forecasting}.

Another aspect of modeling human behavior is modeling its intrinsic decisions. There have been attempts to model the rationality \cite{reddy2018you} and legibility \cite{dragan2013legibility} in human-robot interaction. In imitation learning, the objective is to learn a policy to imitate a set of human demonstrations. Behavioral cloning \cite{ross2011reduction} is one way to learn such demonstrations from data. However, such supervised learning techniques have proven to be less successful in applications such as modeling multi-agent traffic due to compounding errors and not taking into account multi-agent interactions \cite{bhattacharyya2020modeling}. Techniques such as inverse reinforcement learning \cite{abbeel2004apprenticeship, gombolay2018human} and inverse reward design \cite{hadfield2017inverse} attempt to directly model the underlying reward function of humans. However, it is not clear how to incorporate traffic rules and road geometry into these models.

The intelligent driver model (IDM) is a widely used rule-based dynamics model used for human driving behavior \cite{treiber2000congested}. It can be used to drive a vehicle at a desired speed in a specific lane while maintaining a minimum spacing with the leading vehicle. Augmenting IDM with MOBIL~\cite{kesting2007general}, another rule-based model, can be used to switch lanes. Even though, by construction, these models are guaranteed to avoid collisions, they do not (1) consider interactions with other vehicles beside the leading vehicle, (2) reflect natural human driving styles as parameters are arbitrarily set by the model user, and (3) account for individual driving behavior because these deterministic models can only have a single set of scalar-valued parameters.

To mitigate some of the limitations of rule-based driving models, especially the multiagent interactions, generative adversarial imitation learning has been used for highway driver modeling \cite{bhattacharyya2020modeling}. Since it is a purely data-driven technique, it exhibits some undesirable behaviors such as collisions. A recent approach has proposed a way to learn distributional parameters of an IDM from data~\cite{bhattacharyya2020online} to guarantee collision-free driving.
\section{Rule-Based Driver Models} \label{sec:drivermodeling}
\subsection{Intelligent Driver Model and Extensions}
The IDM~\cite{treiber2000congested} is a parametric rule-based car-following model that balances two forces: the desire to achieve free speed if there were no vehicle in front, and the need to maintain safe separation with the vehicle in front.
The IDM is guaranteed to be collision-free by construction.
The inputs to the model are the vehicle's current speed $v(t)$ at time $t$, relative speed $r(t)$ with respect to the leading vehicle, and distance headway $d(t)$.
The model then outputs an acceleration according to
\begin{equation}
   a_{\mathrm{IDM}} = a_{\mathrm{max}}\Bigg( 1-\bigg(\frac{v(t)}{v_{\mathrm{des}}} \bigg)^4 - \bigg( \frac{d_{\mathrm{des}}}{d(t)} \bigg)^2 \Bigg) \text{,}
   \label{eqn:a_idm}
\end{equation}
where the desired distance is
\begin{equation}
    d_{\mathrm{des}} = d_{\mathrm{min}} + \tau .v(t) - \frac{v(t).r(t)}{2\sqrt{a_{\mathrm{max}}.b_{\mathrm{pref}}}} \text{.}
    \label{eqn:d_des}
\end{equation}

The model has several parameters that determine the acceleration output based on the scene information. Here, $v_{\mathrm{des}}$ refers to the \raunak{free flow speed}, $d_{\mathrm{min}}$ refers to the minimum allowable separation between the ego and leader vehicle, $\tau$ refers to the minimum time separation allowable between ego and leader vehicle, and $a_{\mathrm{max}}$ and $b_{\mathrm{pref}}$ refer to the limits on the acceleration and deceleration, respectively. Though the collision-free motion of a vehicle can be simulated by arbitrarily setting some parameter values, the driving behavior is not necessarily realistic. Therefore, in this paper, we learn the parameters from real human driver demonstrations.

There are various extensions of the original IDM.
The Enhanced IDM incorporates a slight modification that prevents the model from ``over-reacting'' when another vehicle cuts in front of it \cite{Kesting2010}.
The Foresighted Driver Model modifies the output of the IDM based on factors such as upcoming curvature in the road~\cite{Eggert}.
\citet{Liebner} incorporate a spatially varying velocity profile within the IDM to account for variation in different types of maneuvers through intersections.
% Various proposed stochastic extensions of the IDM are functionally equivalent to ours, although (to the best of our knowledge) none of these is used for online estimation of ``stochasticity'' parameters for individual drivers.
    \citet{Hoermann2017} use a stochastic IDM model with fixed-variance additive Gaussian white noise. % They use a particle filter to estimate IDM parameters online. The key difference between their approach and ours is that we model stochasticity on the level of individual drivers.
    \citet{Schulz} use a similar model that also incorporates context-dependent upper and lower bounds on acceleration \cite{Schulz,Schulz2018}. 
    % In their model, the predicted acceleration command is drawn from a fixed-covariance Gaussian distribution and clipped at the lower and upper bounds. % All model parameters are set heuristically offline.

%%% Parameter Estimation

%% Offline
% Parametric (IDM and friends)
% - Constrained nonlinear optimization \cite{lefevre2014comparison}
% - Levenberg-Marquardt algorithm \cite{morton2017analysis}
% - Spatially varying velocity profile from K-means clustering is embedded in IDM model whose other parameters are set manually (heuristically) \cite{Liebner}
% - Heuristic \cite{Shulz}, \cite{Schulz2018}
% - Recommended IDM values \cite{treiber2017intelligent}
Many approaches in the literature estimate IDM model parameters offline.
    \citet{lefevre2014comparison} use constrained nonlinear optimization.
    \citet{morton2016analysis} use the Levenberg-Marquardt algorithm.
    Some approaches select the parameters heuristically \cite{Schulz,Schulz2018}.
    In fact, ``recommended'' parameter values have been published for the IDM \cite{treiber2017intelligent}.

Offline estimation is also used for selecting parameter values in black-box driver models.
    \citet{lefevre2014comparison} use Expectation Maximization (EM) to train a Gaussian mixture model (GMM), and the Levenberg-Marquardt algorithm to train a neural network (NN).
    \citet{morton2016analysis} use gradient-based optimization to train various feedforward and recurrent neural network models.
    \citet{kuefler2017imitating} use Generative Adversarial Imitation Learning (GAIL) to train a recurrent neural network.
    
Some approaches estimate driver model parameters online.
    In Multi-Policy Decision-Making, the parameters of several hand-crafted control policies are estimated online with Bayesian Changepoint Estimation and Maximum-likelihood estimation \cite{Galceran2017a}.
    \citet{Sadigh2018} use online active information gathering to estimate the parameters of a human driver's reward function.

Several online estimation approaches are used for IDM in particular.
    \citet{monteil2015real} use an Extended Kalman filter.
    Examples of particle filters used with IDM parameters include approximate online POMDP solvers \cite{Sunberg2017a} and fully probabilistic scene prediction algorithms \cite{Hoermann2017}. The online parameter estimation approach of \citet{Buyer2019} is similar to ours, although they use a different IDM extension and do not use their model for forward simulation of traffic scenes \cite{Buyer2019}. 
    None of the above models explicitly estimate ``stochasticity'' parameters for individual drivers.

\subsection{Cooperative-IDM for Merging}
We extend the Cooperative Intelligent Driver Model (C-IDM) as originally proposed by \citet{bouton2019cooperation}. In addition to the IDM parameters, C-IDM includes a cooperation parameter ${c \in [0,1]}$, which controls the level of cooperation to the merging vehicle. With ${c=1}$, the driver slows down to yield to the merging vehicle and with ${c=0}$, the driver completely ignores the merging vehicle until it traverses the merge point, after which it follows IDM. 
\raunak{C-IDM relies on estimating the time to reach the merge point for the car on the main lane and the car on the merge lane to decide whether the merging vehicles should be considered.
While C-IDM was originally proposed to work with only one vehicle merging into a main lane, we extend it to work with multiple merging vehicles.
The ego vehicle performs a search over the vehicles in the merging lane to select the one with the closest time to the merge point value.
In this work, we used a simple constant velocity model to estimate the time to merge (TTM). 
A more sophisticated prediction model can be used to have more realistic estimates of the time to merge.
Based on the cooperation parameter, two cases are considered:
\begin{itemize}
    \item If $\text{TTM}_{ego} < c \times \text{TTM}_{other}$, the ego vehicle follows IDM by considering the projection of the other vehicle onto its own lane
    \item If $\text{TTM}_{ego} \geq c \times \text{TTM}_{other}$, the ego vehicle ignores the vehicle on the merging lane.
\end{itemize}
}

\section{Driver Modeling using Particle Filtering}
This section defines the problem of learning human driving models from demonstrations and discusses our methodology of using particle filtering for online parameter estimation for rule-based driver models.

\subsection{Problem Definition}
We are given a batch of trajectories $\mathbf{y}_{1:T}$ 
% $\in \mathbb{R}^{D \times T}$ 
of human driving demonstrations over a time horizon $T$. 
We assume that human driving follows a dynamical system with state 
% $\mathbf{x} \in \mathbb{R}^D$ 
$\mathbf{y}$
that evolves according to the following equation:
% \begin{equation}
% \begin{split}
%     \mathbf{x}_{t+1} &= f_{\theta_t}(\mathbf{x}_t,\mathbf{w}_t) \\
%     \mathbf{y}_t &= g_{\theta_t}(\mathbf{x}_t,\mathbf{v}_t) \text{,}
% \end{split}
% \end{equation}
\begin{equation}
    \mathbf{y}_{t+1} = f_{\theta_t}(\mathbf{y}_t,\mathbf{w}_t) \text{,}
\end{equation}
where $\mathbf{w}_t$ is the process noise. 
% and $\mathbf{v}_t$ is the observation noise at time $t$.
We assume that we are provided a class of parametrized models $f_\theta(x,t)$ which represents the underlying rule-based driver model.
% and $g_\theta(x,t)$ 
% and observation processes, respectively. 
% \red{Connect to how other fields such as IRL or System ID view human demonstrations. For example, System ID wants to find point estimates of the parameters for purposes of control and state estimation as opposed to finding distributions over the parameters like we do. And IRL views the underlying demonstration as an MDP, which I think is a special case of dynamical systems?}

Our goal is to use the demonstrated driving trajectories to learn a distribution over the parameter vector $\theta$ of the function $f$.
% and $g$.
This distribution represents the variation in the possible human behaviors.

\begin{algorithm}
  \caption{Driver model parameter estimation using particle filtering}
  \label{algo:pfidm}
  \begin{algorithmic}
    \STATE {\bfseries Input:} Demonstration trajectories of length $T$, Starting scene with $K$ humans, Initial set of particle sets $\{\Theta_1, \Theta_2, \dotsc, \Theta_K\}$

    \FOR[humans] {$k \gets 1,2, \dotsc, K$}
    
    \FOR[time-steps]{$t \gets 0, 1, \dotsc, T$}
    
    \STATE $\mathbf{y}_k^{(t+1)} \gets$ ground truth observation of the $k$th human at $t+1$
    %\STATE $\Theta_k \gets \{ \}$
    \FOR[particles]{$i \gets 1, 2, \dotsc, I_k$}
    \STATE $\theta_i \gets$ random particle in $\Theta_k$
    \STATE $\mathbf{y}_{k,i}^{(t+1)} \sim \text{Observation evolution}$ \COMMENT{Sampled next observation for $k$th human using the $i$th particle}
    \STATE $w_i \gets O\big( \mathbf{y}_k^{(t+1)}\mid \mathbf{y}_{k,i}^{(t+1)} \big)$ \COMMENT{Probability density of true next observation given sampled next observation}
    \ENDFOR
    \STATE $\Theta_k \gets$ Obtain $I_k$ samples from $\Theta_k$ according to $[w_1, w_2, \dotsc, w_{I_k}]$ \COMMENT{Resampling}
    \STATE Step trajectory forward by one time-step
    \ENDFOR
    \ENDFOR
    \STATE Combine particles obtained from all humans and use as prior for next epoch
  \end{algorithmic}
\end{algorithm}

\subsection{Latent Parameter Estimation}
\begin{figure}[t]

\centering
    \includegraphics[width=\columnwidth]{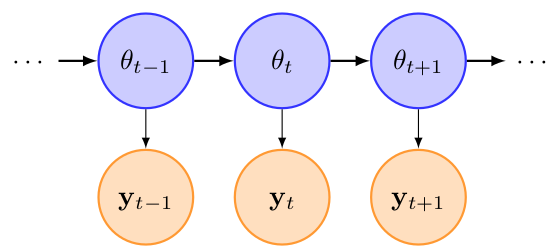}

\caption{Hidden Markov model showing the driver modeling problem in the state estimation framework. $\theta_t$ and $\mathbf{y}_t$ are the (hidden) parameters of the driver model and the vehicle position at time $t$, respectively. The objective is to learn a distribution over the latent parameters using data from multiple human demonstrations.
    }
    \label{fig:hmm}
\end{figure}

We view the problem of finding parameter distributions from the lens of state estimation. 
The parameters of the model evolve according to a hidden Markov model where the transition distribution is governed by $p(\theta_{t+1} \mid \theta_t)$ and the observation distribution is governed by $p(\mathbf{y}_t \mid \theta_t)$ as shown in Figure~\ref{fig:hmm}. 
Our goal is to find the posterior distribution $p(\theta_T \mid \mathbf{y}_{1:T})$.

% https://texample.net/tikz/examples/kalman-filter/

The distribution over parameters can be written as
\begin{equation}
    p(\theta_T \mid \mathbf{y}_1,\mathbf{y}_2,\dots,\mathbf{y}_T) \text{,}
    \label{eqn:inference}
\end{equation}
\noindent where ($\mathbf{y}_1,\mathbf{y}_2,\dots,\mathbf{y}_T$) denotes a sequence of observations from demonstration data. This inference problem can be solved using recursive Bayesian estimation, where the recursive update equation is given by
\begin{equation}
    p(\theta_T \mid \mathbf{y}_{1:t}) = \frac{p(\mathbf{y}_t \mid \theta_T)p(\theta_T \mid \mathbf{y}_{1:t-1})}
    {\int_{\theta_T} p(\mathbf{y}_t \mid \theta_T)p(\theta_T \mid \mathbf{y}_{1:t-1}) \mathrm{d}\theta_T} \text{.}
    \label{eqn:recursive_state_estimation}
\end{equation}

The partition function (the denominator) in (\ref{eqn:recursive_state_estimation}) cannot be evaluated analytically for general nonlinear distributions.
Rather than imposing restrictive assumptions on the form of the distribution, we use particle filtering~\cite{liu1998sequential,thrun2002particle} to approximately solve the inference problem. 
% Further, different humans have different behaviors which cannot be represented as a parametrized distribution. 
A particle filter approximates a continuous probability distribution with a collection of sampled particles. The parameters also include the inherent stochasticity in human behavior, and the particles represent our uncertainty over this stochasticity.

To model human driving behavior, which is inherently stochastic (given the  same scene, a human driver may not always take the same resulting action), we inject stochasticity. 
We assume that the output acceleration is distributed according to 
\begin{equation}
    a \sim \mathcal{N}( a_{\mathrm{IDM}},\sigma_{\mathrm{IDM}}) \text{,}
    \label{eqn:errorable_idm}
\end{equation}
where $a_{\mathrm{IDM}}$ and $\sigma_{\mathrm{IDM}}$ represent the mean and variance, respectively, of a Gaussian distribution. The mean $a_{\mathrm{IDM}}$ is the acceleration output, and $\sigma_{\mathrm{IDM}}$ is a new model parameter representing execution noise.
Assuming the dynamics
\begin{equation}
    y_{t+1} = y_t + \frac{1}{2}a \Delta t^2 \text{,}
    \label{eqn:dynamics}
\end{equation}
where $y$ is the position and $\Delta t$ is the unit-time, we obtain the new position distributed according to
\begin{equation}
    y_{t+1} \sim \mathcal{N} (y_t + \frac{1}{2}a_{\mathrm{IDM}} \Delta t^2,\sigma_{\mathrm{IDM}} \Delta t^2) \text{.}
    \label{eqn:position_update}
\end{equation}

We perform particle filtering over the trajectory provided by one human demonstrator to reach a distribution over the parameters for that particular demonstrator. We then mix the distributions obtained from multiple human demonstrators. 
To maintain the simplicity of our approach, in this work, we combine the sample-based representation of the distributions obtained from different demonstration trajectories. This constitutes one epoch of our approach. For the subsequent epoch, the distribution learned from the previous epoch is used as a prior from which the initial particle set is sampled. Our algorithm is shown in Algorithm~\ref{algo:pfidm}.

\begin{figure*}[t]
\includegraphics[scale=0.33]{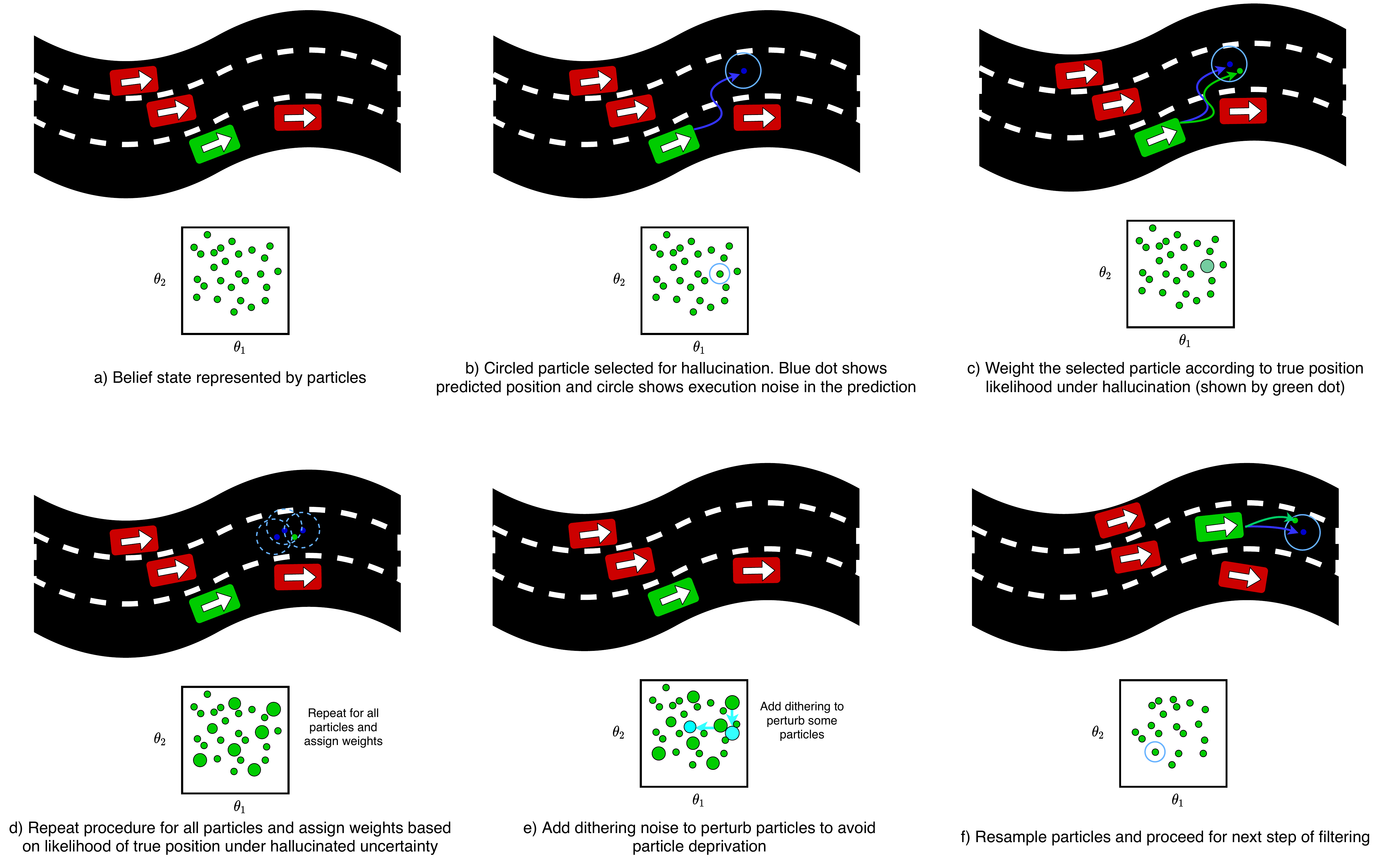}
\caption{The particle filtering process to learn a distribution over the parameters of the underlying rule-based driver model from ground truth demonstration data. The vehicle of interest (green) interacts with the surrounding vehicles (red). The blue trajectories show hallucinations carried out by different particles and the green trajectory shows the ground truth. The ground truth position likelihood under the distribution over hallucinated position is used to weight and resample the particles.}
\label{fig:carfiltering}
\end{figure*}

Figure~\ref{fig:carfiltering} illustrates the particle filtering procedure in our driver modeling case study. 
    For the purposes of illustration, we assume that the parameter space of the driver model is 2 dimensional (the actual parameter space is 8 dimensional). 
        First, a set of particles is sampled from a uniform distribution. 
            Each particle is then used to hallucinate the vehicle one step forward. 
                The hallucinated position is used as the mean of a bivariate Gaussian distribution whose covariance is governed by the stochasticity parameters. 
                    The mean position and the uncertainty are shown by the blue dot and circle respectively in panel b). 
                        Subsequently, the particle is weighted according the likelihood of the ground truth position under the bivariate Gaussian distribution. \raunak{This ground truth position is obtained from the demonstration trajectory provided by the real world human driving dataset.}
                            A similar procedure is carried out to assign weights to all the particles in the particle set resulting in a weighted particle set as shown in panel d). 
                                To counter the particle deprivation problem, a small amount of noise is added to the particle set. 
                                    Finally, the vehicles are moved one step forward according to the ground truth trajectory, and the particle set is resampled according to the weighted set obtained in panel e). 
\section{Experiments}
In this section, we demonstrate our methodology on two driver modeling tasks using three real-world datasets.

\subsection{Experiments on Highway Driving}
We evaluate the performance of our model on demonstration data from two real-world datasets, namely the Next-Generation Simulation (NGSIM) for US Highway 101~\cite{colyar2007us} which provides driving data collected at \SI{10}{\hertz} and the Highway Drone Dataset (HighD)~\cite{highDdataset} which provides driving data from German highways recorded at \SI{25}{\hertz} using a drone.
    \raunak{Traffic density in the NGSIM dataset transitions from uncongested to full congestion and exhibits a high degree of vehicle interaction as vehicles merge on and off the highway and must navigate in congested flow.
        On the other hand, the HighD dataset has relatively free flow traffic.
            The trajectories were smoothed using an extended Kalman filter on a bicycle model and projected to lanes using centerlines extracted from the NGSIM CAD files.
            }

We benchmark our approach against representative rule-based and black-box models as well as constant velocity and constant acceleration baselines.
    The code for all the experiments is publicly available at our code base.\footnote{\href{https://github.com/sisl/ngsim_env/tree/idm_pf_NGSIM}{https://github.com/sisl/ngsim\_env/tree/idm\_pf\_NGSIM}}

To estimate the parameters of the IDM using our filtering approach as per~\cref{algo:pfidm}, the particles are initially sampled from a uniform distribution discretized into a grid with resolution of \SI{0.5}{\meter/\second} for the desired velocity parameter ($v_{\mathrm{des}}$) and $0.1$ for the stochasticity parameter ($\sigma_{\mathrm{IDM}}$). 
At the dithering stage (to avoid particle deprivation), we add noise sampled from a discrete uniform distribution with ${v_{\mathrm{des}} \in \{-0.5,0,0.5\}}$ and ${\sigma_{\mathrm{IDM}} \in \{-0.1,0,0.1\}}$.
These values are chosen to preserve the discretization present in the initial sampling of particles.
The time taken for filtering to converge in a 20 vehicle scenario over a \SI{5}{\second} duration was \SI{30}{\second} on an Intel Core i9-9900K eight-core processor.

\begin{figure}
    \centering
    \includegraphics[width=\columnwidth]{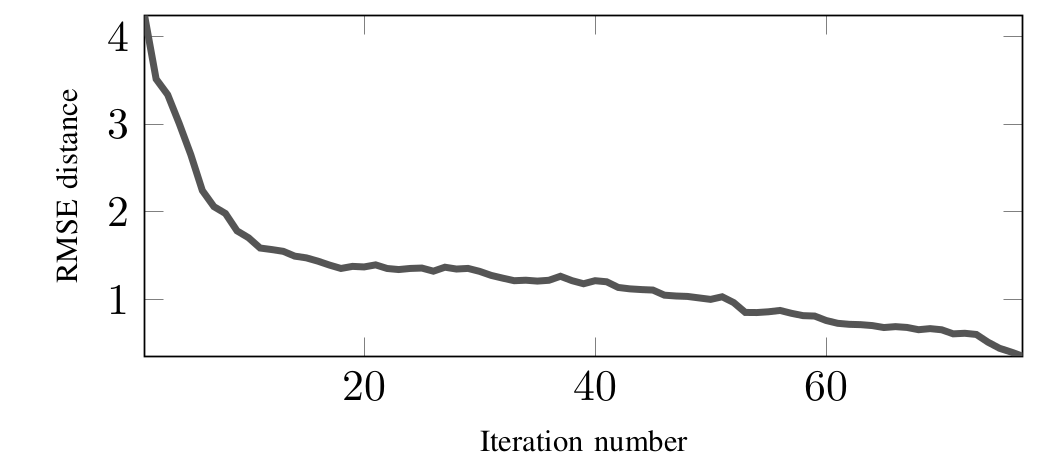}
    % \begin{tikzpicture}
    % \begin{axis}
    % [enlargelimits=0.0,
    % height = 0.5\columnwidth,
    % width = \columnwidth,
    % xlabel=\scriptsize Iteration number,
    % ylabel=\scriptsize RMSE distance,
    % ]
    %     \addplot[colorD, ultra thick] 
    %     table[x=iter, y=val, col sep=comma]{figs/pf_prog.csv};
    % \end{axis}
    % \end{tikzpicture}
    \caption{\footnotesize RMSE distance from final particle over particle set at every iteration averaged over all the vehicles. The particle set converges to the final particle with the progress of filtering.}
    \label{fig:filtering_progress}
    \setlength{\belowcaptionskip}{-15pt}
\end{figure}

To assess the convergence of the particle filtering approach,~\cref{fig:filtering_progress} shows the root mean squared distance from the mean of the final particle distribution over the set of particles at every iteration.
The particles converge as more demonstration data is shown to the filtering algorithm.

%*********Plot:Filtering progress*****

\Cref{fig:carwise_mean_particle} shows the mean particle after the filtering process for a subset of vehicles from both the NGSIM and the HighD datasets. 
The HighD vehicles have a higher desired velocity ($v_\mathrm{des}$) parameter on average, reflecting the fact that vehicles drive faster on German highways.

\input{figs/carwise_mean.tex}

Experiments are conducted on a set of thirty scenarios (fifteen scenarios randomly sampled from each dataset). Twenty vehicles in each scenario are randomly selected as target vehicles.
For each scenario and each model, predicted trajectories are generated by forward simulation of this set of target vehicles over a \SI{5}{\second} time horizon, where the target vehicles are controlled by the driver model defined by the parameters estimated using particle filtering.

We use root mean square error (RMSE) of the position and velocity to measure ``closeness'' of a predicted trajectory to the corresponding ground-truth trajectory.
% \Cref{fig:rmse} shows the RMSE over time for an example scenario with 20 vehicles over a \SI{5}{\second} duration from the NGSIM dataset.

While RMSE measures prediction accuracy at the level of individual vehicles by comparing the obtained trajectories against ground truth from the demonstration trajectories, we also wish to quantify how ``safely'' each model drives. To this end, we count the number of ``undesirable events'' (collision, going off the road, and hard braking) that occur in each scene prediction.
% ~\cref{fig:undes_cumsum} shows the cumulative number of undesirable driving instances for 20 vehicles over a $5$ \si{s} duration in a congested traffic scenario in NGSIM. 

To benchmark the performance of our approach, we compare the driving behavior obtained by our model against that obtained by five other models. % Predicted trajectories for each model are generated by forward simulation of the entire traffic scene, where the target vehicle is controlled by the model and all other vehicles follow their ground truth trajectories.
The first benchmark model is IDM with the ``default'' parameter values recommended in ~\cite{treiber2017intelligent}: $v_{\mathrm{des}} = \SI{30}{\meter/\second}$, $\tau=\SI{1.0}{\second}$, $d_{\mathrm{min}}=\SI{2}{\meter}$, $a_{\mathrm{max}} = \SI{3}{\meter/\second^2}$, and $b_{\mathrm{pref}} = \SI{2}{\meter/\second^2}$.
Our second (also rule-based) benchmark model is the IDM with parameters obtained by offline estimation using non-linear least squares~\cite{morton2016analysis}.
The associated parameter values are $v_{\mathrm{des}} = \SI{17.837}{\meter/\second}$, $\tau= \SI{0.918}{\second}$, $d_{\mathrm{min}}=\SI{5.249}{\meter}$, $a_{\mathrm{max}} = \SI{0.758}{\meter/\second^2}$, and $b_{\mathrm{pref}} = \SI{3.811}{\meter/\second^2}$.
Our third benchmark model is a recurrent network trained with Generative Adversarial Imitation Learning (GAIL)~\cite{bhattacharyya2018multi}.
% GAIL models the driving task as a sequential decision making problem and searches for policies that map road scene to actions.
% GAIL learns driving policies directly from driving demonstration data according to minimizing distance between expert policy and driving policy.
% Unfortunately, GAIL does not reliably give collision-free driving behavior.
We also baseline our method against constant velocity (vehicles continue driving at the same speed that they start with at the beginning of the simulation) and constant acceleration (vehicles accelerating at $\SI{1}{\meter/\second^2}$) models.

\input{figs/rmse.tex}

RMSE results for an example scenario with 20 vehicles over a \SI{5}{\second} duration from the NGSIM dataset are shown in~\cref{fig:rmse}.
We observe that our method provides driving trajectories that are closer to the ground truth as compared to those generated by IDM with default parameter values and those generated by GAIL driving policies.
    We see that the RMSE in both position and velocity averaged over the set of vehicles is lowest for all timesteps using our driving model.

Further experiments on both NGSIM and HighD datasets are reported in~\cref{table:comp}. These results are generated using 15 randomly sampled scenarios from both the HighD and NGSIM datasets. 
Every scenario is such that there is a set of 20 vehicles driving over a \SI{5}{\second} horizon which translates to 50 timesteps for NGSIM and 125 timesteps for HighD.
We see that while our method outperforms other methods, it performs worse than the constant velocity baseline for the HighD dataset. 
One possible reason may be the default values for the parameters that govern the interaction between vehicles, i.e. minimum allowed separation $d_\mathrm{min}$ and minimum timegap $\tau$.
Including these parameters within the filtering process will allow finer grained driver modeling and is an interesting direction for future work.

\begin{table*}
  \centering
    \caption{\footnotesize Experiments over 15 randomly selected scenarios for both NGSIM and HighD each with 20 vehicles driving for a $5$ \si{s} duration. The results show the RMSE values for position and velocity at the end of $5$ \si{s}. Cumulative number of collisions at the end of the horizon are also reported.}
    \begin{tabular}{@{}llcccccc@{}}
    \toprule
     &&\multicolumn{5}{c}{\bf Models}  \\ \cmidrule{3-8}
      {\bf Metrics}&{\bf Dataset} & $\mathrm{IDM}_{\theta}$ (ours) & Default \cite{treiber2017intelligent} & GAIL \cite{bhattacharyya2018multi} & Const. Speed & Const. Acc. & Non-Linear Fit~\cite{morton2016analysis}\\
      \midrule
    %   \parbox[t]{2mm}{\multirow{9}{*}{\rotatebox[origin=c]{90}{\bf Metrics}}}
      \parbox[t]{2.2cm}{\multirow{2}{*}{Position RMSE}}  
      &NGSIM & 5.90 $\pm$ 1.98 &
      27.78 $\pm$ 5.40 &
      10.42 $\pm$ 3.73 & 
      6.24 $\pm$ 2.02 &
      12.64 $\pm$ 4.70 &
      7.34 $\pm$ 4.55\\
     &HighD & 8.02 $\pm$ 3.34 &
     18.30 $\pm$ 9.03 & 
     13.63 $\pm$ 3.92 &
     2.42 $\pm$ 1.64 & 
     11.01 $\pm$ 1.92 &
     35.13 $\pm$ 7.21\\
      \midrule
     \parbox[t]{2.2cm}{\multirow{2}{*}{Velocity RMSE}}
     &NGSIM & 2.12 $\pm$ 0.79 &
     10.72 $\pm$ 2.36 &
     3.52 $\pm$ 1.28 &
     2.22 $\pm$ 0.82 &
     5.03 $\pm$ 1.78 &
     2.69 $\pm$ 1.77\\
       &HighD & 2.14 $\pm$ 0.65 &
       4.59 $\pm$ 2.46 &
       2.94 $\pm$ 0.93 &
       0.94 $\pm$ 0.57 &
       4.39 $\pm$ 0.61 &
       10.05 $\pm$ 2.07\\
      \midrule
      \parbox[t]{2.2cm}{\multirow{2}{*}{Number of collisions}}
      &NGSIM & 0 $\pm$ 0 &
      0 $\pm$ 0 &
      53 $\pm$ 11 &
      113 $\pm$ 18 &
      119 $\pm$ 16 &
      0 $\pm$ 0\\
       &HighD & 0 $\pm$ 0 &
       0 $\pm$ 0 & 
       15 $\pm$ 4 &
       0 $\pm$ 0 &
       27 $\pm$ 3 &
       0 $\pm$ 0\\
      \bottomrule
    \end{tabular}
  \label{table:comp}
\end{table*}

\input{figs/undesirable.tex}

% While RMSE measures the driving performance at the level of individual vehicles by comparing the obtained trajectories against ground truth from the demonstration trajectories, the safety of the driver model is assessed in a congested driving scenario.
The cumulative number of undesirable driving instances for 20 vehicles over a \SI{5}{\second} duration in a congested traffic scenario from the NGSIM dataset is shown in ~\cref{fig:undes_cumsum}.
    The cumulative number of undesirable driving instances keep growing with time for the data-driven benchmark in~\cref{fig:undes_cumsum}.
This reflects the fact that GAIL does not provide guarantees on safety.
As expected, the IDM based models, including ours and the two rule-based benchmarks, do not show any collisions because the IDM is collision-free by default.
The constant velocity and constant acceleration baselines also do not provide collision-free trajectories because they are not reacting to the vehicle in front of them but merely driving with constant velocity and acceleration, respectively.

Cumulative number of collisions for all vehicles over the duration of the trajectory are also reported in~\cref{table:comp}.
We observe that the constant velocity baseline suffers from no collisions in the HighD dataset.
This is because the dataset is not as congested as the NGSIM dataset and hence vehicles start with sufficient distance headway and relative velocity to avoid collisions.
However, the constant acceleration does result in some collisions whenever a faster vehicle starts out behind a slower vehicle.
The data-driven benchmark also results in some collisions (fewer than NGSIM due to larger separation between vehicles).
As expected, congested scenarios present a challenge for the benchmark models.
\subsection{Experiments on Merging}
\begin{figure*}[t]
    \centering
    \includegraphics[width=\textwidth]{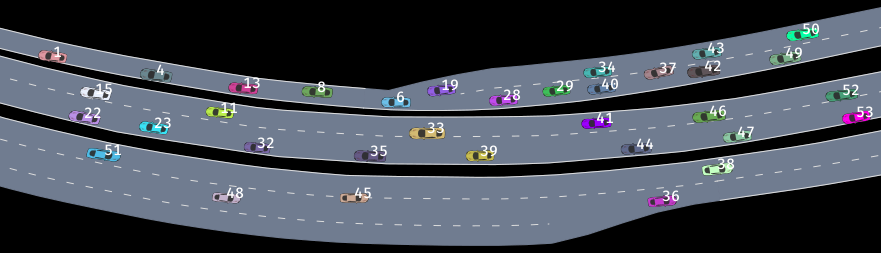}
    % \includesvg[width=\textwidth]{dataset.svg}
    \caption{The Interaction Dataset~\cite{interactiondataset} contains real-world driving demonstrations. In this case study, we model highway merging from demonstrations. This figure shows one time snapshot from an example scenario from the dataset. Here, vehicles 19, 34, 43 and 50 are attempting to merge into the main lane. Using particle filtering, we learn a distribution over the parameters governing the rule-based Cooperative Intelligent Driver Model~\cite{bouton2019cooperation} to best capture the demonstrated driving behavior.}
    \label{fig:dataset}
\end{figure*}

We demonstrate our proposed methodology in the case of modeling highway merging behavior. We choose merging as it is a complex problem involving negotiation between multiple drivers and implicitly inferring the intent of these drivers. Prior approaches to modeling merging behavior have been in the context of planning. \citet{bouton2019cooperation} demonstrate the use of hierarchical reinforcement learning to enable an ego vehicle to safely merge in dense traffic. Other approaches to merging include online planning~\cite{ward2017probabilistic,hubmann2018belief,schmerling2018multimodal} and game theoretic methods~\cite{sadigh2018planning,fisac2019hierarchical}, both of which suffer from the lack of ability to scale.
    The code for our highway merging experiments on the demonstration data from the Interaction Dataset is publicly available at our code base.\footnote{\href{https://github.com/sisl/AutomotiveInteraction.jl}{https://github.com/sisl/AutomotiveInteraction.jl}}

The Interaction Dataset~\cite{interactiondataset} contains interactive driving scenarios from different countries. For this work, we focus on a merging scenario. Figure~\ref{fig:dataset} shows the map and the vehicles at one time snapshot in our driving simulation platform. There are two merge lanes from both directions. Further, the merge lanes allow lane changes into the main lanes. Our goal is to obtain a driver model that performs merging like humans do based on the demonstrations provided in the merging scenario. Further, we want to use these driver models to generate novel scenarios of interest.

We take demonstration trajectories from 15 scenarios, which average \SI{50}{\second} in duration involving a total of 87 vehicles. Particle filtering is used to find a distribution over the parameters of the underlying Cooperative-IDM with stochasticity. Parameters are then sampled from this distribution to generate rollout trajectories. We extract metrics of imitation performance to assess how closely our rollout trajectories match those of the demonstrations. In addition, we extract the number of collisions to assess the quality of the generated driving behavior.

We baseline our models against three types of driver models. The first baseline model is vanilla IDM that uses parameter values as obtained by heuristics~\cite{treiber2000congested}, representing a purely rule-based model. The second baseline model is C-IDM, which has a cooperation parameter of 1 in addition to the parameter values for IDM. This represents the most cautious setting of the Cooperative-IDM~\cite{bouton2019cooperation}. The final baseline is LMIDM, which represents the addition of data processing to underlying rules. In LMIDM, we use non-linear least squares to estimate the parameters from data using the Levenberg-Marquardt (LM) algorithm~\cite{more1978levenberg}.

\input{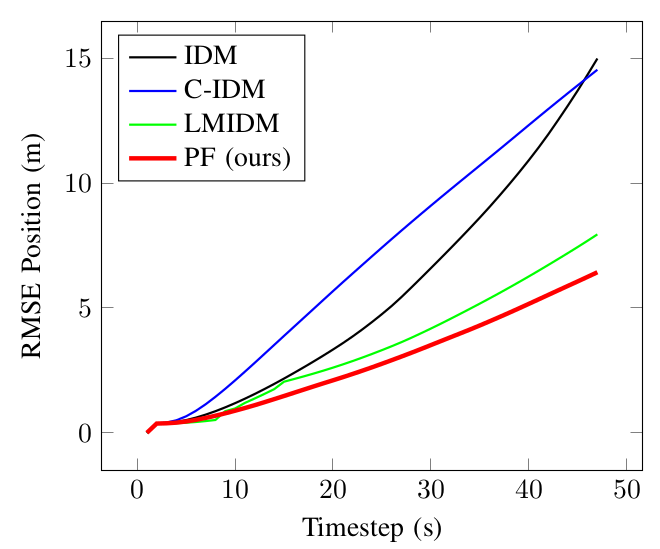}
Figure~\ref{fig:rmsepos} shows the imitation performance in terms of how closely the generated trajectories match the ground truth demonstration trajectories. The particle filtering-based approach to learning parameters of the Cooperative-IDM shows the lowest RMSE value, highlighting its ability to replicate ground truth trajectories.

% https://tex.stackexchange.com/questions/370539/pgfplots-create-bar-plot-with-bars-in-different-colors
% \begin{figure}
%     \centering
    
%     \begin{tikzpicture}[]
%     \begin{axis}[
%         xmin=0,
%         xmax=5,
%         xtick={1,2,3,4},
%         x tick style={draw=none},
%         xticklabels={IDM,CIDM,LMIDM,PF (ours)},
%         ymin=0,
%         ylabel={Collision Fraction},
%         every axis plot/.append style={
%           ybar,
%           bar width=.2,
%           bar shift=0pt,
%           fill
%         }
%       ]
%       \addplot[black]coordinates {(1,0.07659574468085106)};
%       \addplot[blue]coordinates{(2,0.0)};
%       \addplot[green]coordinates{(3,0.01276595744680851)};
%       \addplot[red]coordinates{(4,0.0)};
%     \end{axis}
%   \end{tikzpicture}
 
% \caption{Number of collisions arising out of simulations carried out using driver models. The vanilla IDM based models show collisions since they are unaware of merging. The particle filtering based and most cooperative c-IDM do not show any collisions.}

% \label{fig:collisions}
% \end{figure}

\begin{table}[b]
\begin{center}
 \begin{tabular}{@{}lr@{}} 
 \toprule
 \textbf{Model} & \textbf{Collision Fraction} \\ 
 \midrule
 IDM & 7.66 \\ 
 C-IDM & 0.00 \\
 LMIDM & 1.28 \\
 PF (ours) & 0.00  \\
 \bottomrule
%  5 & 88 & 788 & 6344 \\ [1ex] 
%  \hline
\end{tabular}
\end{center}
\caption{Collision fraction observed in simulations carried out using different driver models. The vanilla IDM-based models show collisions since they are unaware of merging. The particle filtering-based and the most cooperative C-IDM do not show any collisions.}
\label{table:collisions}
\end{table}
Table~\ref{table:collisions} shows the number of collisions between vehicles arising out of the driver model simulations. Since IDM is a lane follower and not designed to perform merges, it shows a high number of collisions. C-IDM is extremely cautious as expected and therefore does not cause vehicle collisions; however, it suffers from poor RMSE performance as seen in Figure~\ref{fig:rmsepos}. The LMIDM-based models result in some collisions and the particle filtering-based models do not show any collisions.
These results show the benefit of interpretable models. Had we seen collisions while using black-box models, we would not have been able to interpret the behavior leading to those collisions.

% \begin{figure}
% \includegraphics[scale=0.65]{density_567.pdf}
% \caption{Speed distribution over all vehicles obtained from different driver models. The particle filtering based models show the closest speed distribution to the ground truth.}
% \label{fig:speeddist}
% \end{figure}

\input{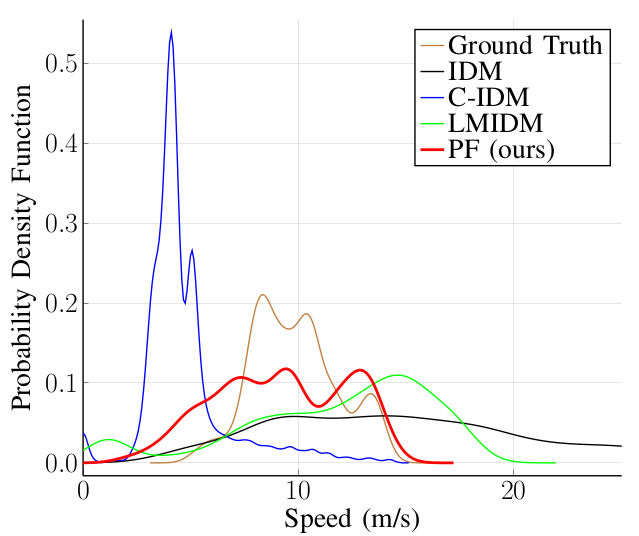}

\Cref{fig:speeddist} shows the speed distribution obtained using driving behavior generated by the different driver models, along with the speed distribution in the ground truth demonstrations. As expected, the speeds obtained by the C-IDM tend toward 0 because of extreme caution. The particle filtering-based driver model shows the closest speed distribution to ground truth, which shows the ability of these models to generate good emergent driving behavior in addition to performing individual trajectory imitation as shown in \cref{fig:rmsepos}.
% LMIDM is hard to explain because of linearity assumptions we don't know what happens. Think about why particle filter does not exactly match the ground truth. This may be because we have not explicitly captured the interaction between vehicles i.e. how a following vehicle slows down because of the vehicle in front slowing down may not be exactly as what IDM does.

\begin{figure}
\renewcommand\arraystretch{1.5}
\setlength\tabcolsep{0pt}
\begin{tabular}{c >{\bfseries}r @{\hspace{0.7em}}c @{\hspace{0.4em}}c @{\hspace{0.7em}}l}
  \multirow{10}{*}{\rotatebox{90}{\parbox{1.1cm}{\bfseries\centering Truth}}} & 
    & \multicolumn{2}{c}{\bfseries Response} & \\
  & & \bfseries Real & \bfseries Synthetic  \\
  & Real & \MyBox{TP}{(60)} & \MyBox{FN}{(45)} \\[2.4em]
  & Synthetic & \MyBox{FP}{(49)} & \MyBox{TN}{(56)} \\
\end{tabular}
\caption{The confusion matrix obtained from the driving Turing test results. Participants were shown videos of real and synthetic driving videos. Real is considered as positive and synthetic as negative for the purposes of the confusion matrix. The numbers in the specific squares represent the number of responses under the categories: true positive (TP), false negative (FN), false positive (FP), and true negative (TN).}
\label{fig:cnfmatrix}
\end{figure}

\raunak{The RMSE results demonstrate how closely our driver models are able to replicate demonstrated trajectories. 
    However, for safety validation in simulation, we need to be able to generate a wide range of driving scenarios.
        We assessed the scenario generation capability of our driver models using a ``driving Turing test.'' 
            We asked 21 human participants to distinguish between driving videos that were a replay from our real world merging dataset (real), and driving videos generated by our driver models (synthetic).
                Correct classifications of the videos as real vs. synthetic resulted in true positives and true negatives, and misclassifications resulted in false positives and false negatives.
}

\raunak{
Figure~\ref{fig:cnfmatrix} shows the confusion matrix obtained from this classification.
    The 60 true positives and 59 true negatives show instances where the human participants correctly classified the video.
        The 49 false positives show instances where the human participants misclassified a synthetic driving video generated by our learned driver models as real-world driving from the dataset. 
            The 45 false negatives represent instances where driving videos showing replays from the driving dataset were misclassifed as being generated from our driver models.
                The accuracy was 55.71\%, indicating that our driver models perform reasonably well in terms of generating realistic driving behavior. Figure~\ref{fig:turing} shows the responses in the form of a table with the columns indicating the scenario number and the rows representing the different human subjects. The black squares indicate misclassification. Almost half the squares are black, which confirms the ability of our driver models to generate realistic human behavior.
}

\input{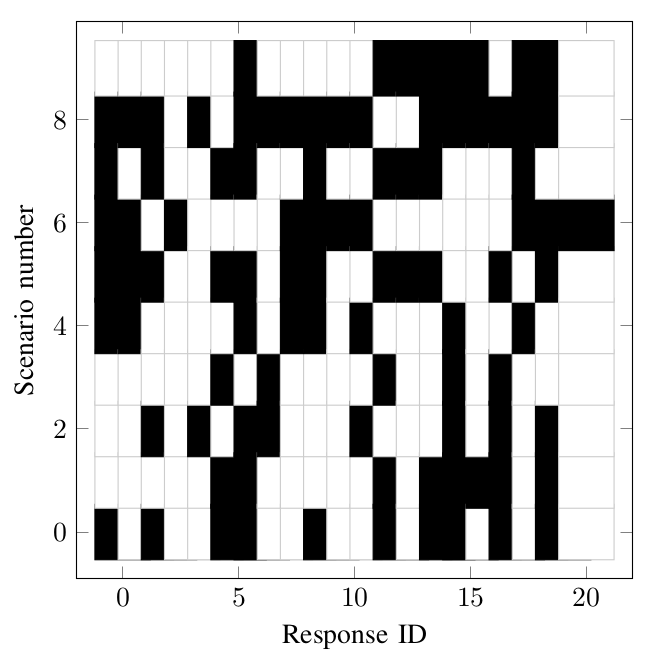}
\section{\liam{Traffic Condition Generation}}

\liam{We briefly discuss scene generation using the stochastic IDM to further explore the effectiveness of rule-based driver models at developing varied traffic conditions for simulation. As previously discussed, simulations must accurately reflect real-world driver behaviors and scenarios to provide meaningful safety validation for autonomous vehicles. Existing real-world datasets such as NGSIM and HighD display a breadth of traffic patterns and vehicle interactions such as merging, congested traffic, and free-flow traffic; however, they ultimately represent brief snapshots of the complex multi-agent interactions that occur across diverse environments and traffic conditions. The ability to generate novel scenes economically in simulation is imperative to ensuring rigorous AV safety validation. In this section, we verify the efficacy of the stochastic IDM at generating varied traffic conditions by testing different parameter combinations to induce \textit{congested} and \textit{free-flow} traffic scenes. We compare aggregate agent behavior arising from the  different traffic conditions; such insights could be leveraged to seed stochastic IDM parameters in future traffic simulations to produce test scenarios for the ego vehicle.}

\begin{figure*}[h]
    \begin{subfigure}[b]{0.19\textwidth}
        % \includesvg[width=\textwidth]{figs/congested/con00.svg}
        \includegraphics[width=\textwidth]{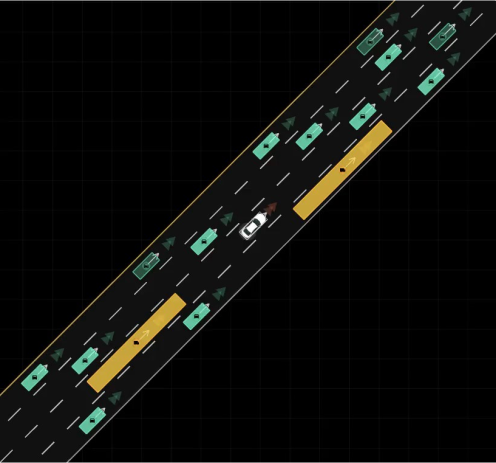}
        \caption{00:00}
        \label{fig:con00}
    \end{subfigure}
    \hfill
    \begin{subfigure}[b]{0.19\textwidth}
        % \includesvg[width=\textwidth]{figs/congested/con15.svg}
        \includegraphics[width=\textwidth]{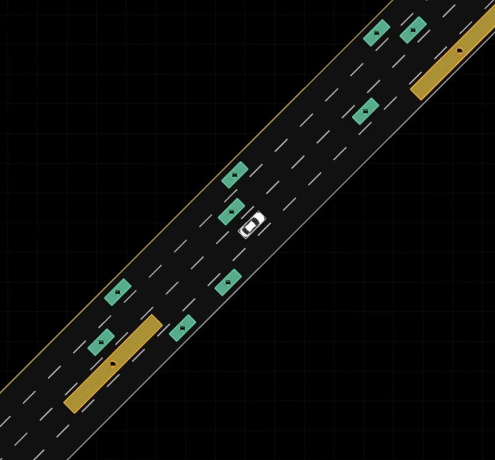}
        \caption{00:15}
        \label{fig:con15}
    \end{subfigure}
    \hfill
    \begin{subfigure}[b]{0.19\textwidth}
        % \includesvg[width=\textwidth]{figs/congested/con30.svg}
        \includegraphics[width=\textwidth]{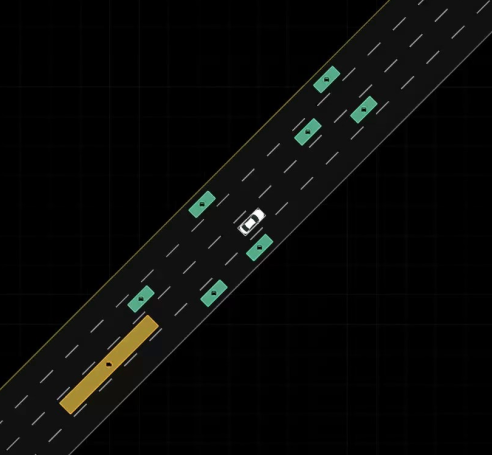}
        \caption{00:30}
        \label{fig:con30}
    \end{subfigure}
    \hfill
    \begin{subfigure}[b]{0.19\textwidth}
        % \includesvg[width=\textwidth]{figs/congested/con45.svg}
        \includegraphics[width=\textwidth]{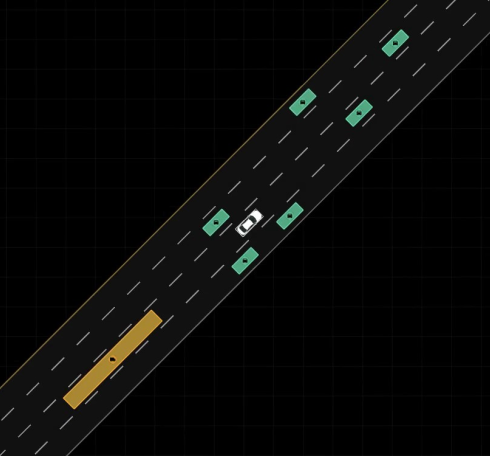}
        \caption{00:45}
        \label{fig:con45}
    \end{subfigure}
    \hfill
    \begin{subfigure}[b]{0.19\textwidth}
        % \includesvg[width=\textwidth]{figs/congested/con60.svg}
        \includegraphics[width=\textwidth]{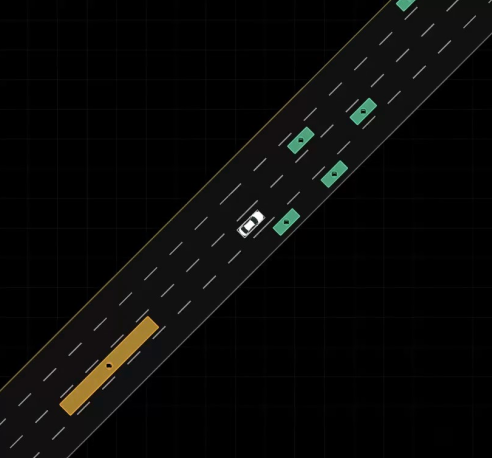}
        \caption{01:00}
        \label{fig:con60}
    \end{subfigure}
\caption{\liam{Scene evolution over 60 seconds for the congested scenario wherein all agent behaviors are defined by parameters drawn from the distributions shown in Table~\ref{table:congested-params}. The agents exhibit a tendency remain clustered together due to relatively small time separation and minimum allowable separation values. The full scene evolution was created using Applied Intuition simulation tools and can be found on the lab video channel at \url{https://youtu.be/ad2148GSs0E}.}}
\label{fig:congested_evolution}
\end{figure*}

\begin{figure*}[h]
    \begin{subfigure}[b]{0.19\textwidth}
        % \includesvg[width=\textwidth]{figs/free_flow/ff00.svg}
        \includegraphics[width=\textwidth]{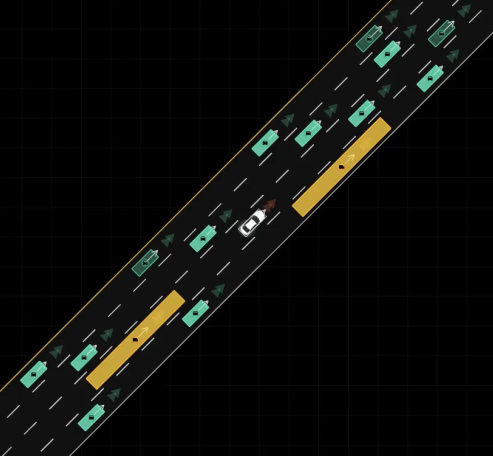}
        \caption{00:00}
        \label{fig:ff00}
    \end{subfigure}
    \hfill
    \begin{subfigure}[b]{0.19\textwidth}
        % \includesvg[width=\textwidth]{figs/free_flow/ff15.svg}
        \includegraphics[width=\textwidth]{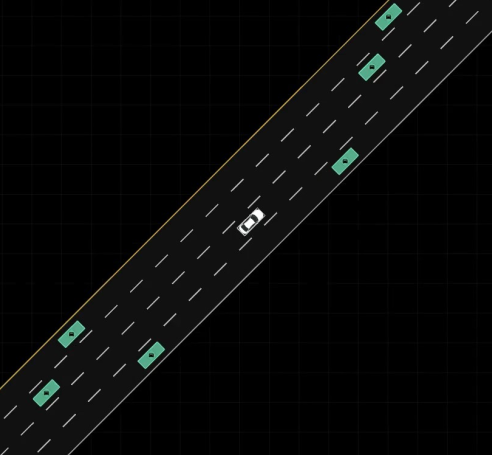}
        \caption{00:15}
        \label{fig:ff15}
    \end{subfigure}
    \hfill
    \begin{subfigure}[b]{0.19\textwidth}
        % \includesvg[width=\textwidth]{figs/free_flow/ff30.svg}
        \includegraphics[width=\textwidth]{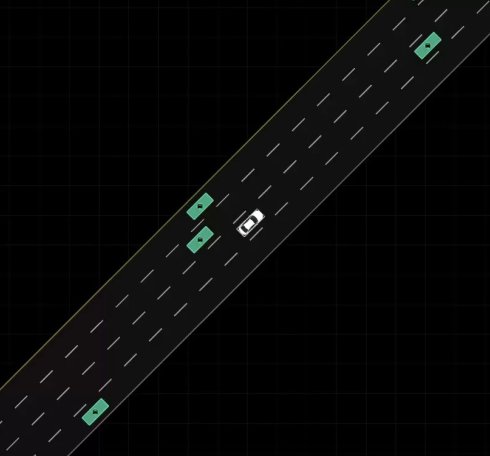}
        \caption{00:30}
        \label{fig:ff30}
    \end{subfigure}
    \hfill
    \begin{subfigure}[b]{0.19\textwidth}
        % \includesvg[width=\textwidth]{figs/free_flow/ff45.svg}
        \includegraphics[width=\textwidth]{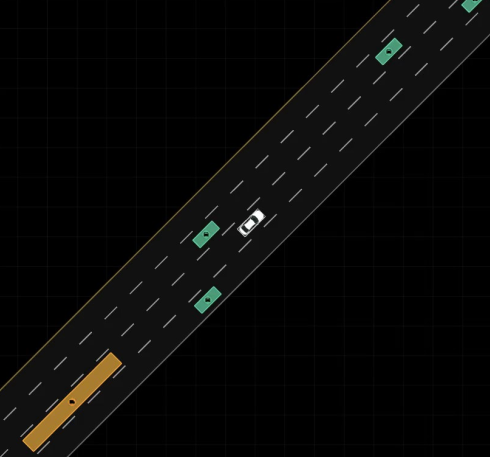}
        \caption{00:45}
        \label{fig:ff45}
    \end{subfigure}
    \hfill
    \begin{subfigure}[b]{0.19\textwidth}
        % \includesvg[width=\textwidth]{figs/free_flow/ff60.svg}
        \includegraphics[width=\textwidth]{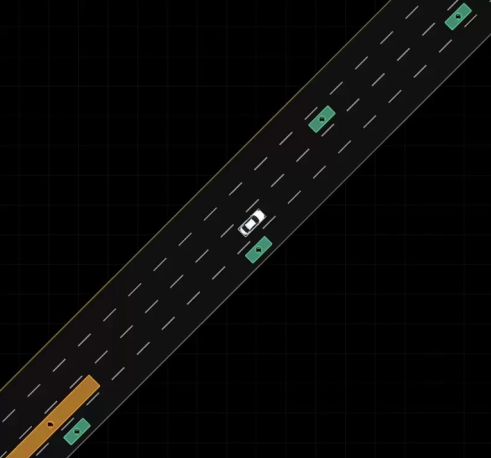}
        \caption{01:00}
        \label{fig:ff60}
    \end{subfigure}
\caption{\liam{Scene evolution over 60 seconds for the free-flow scenario wherein all agent behaviors are defined by parameters drawn from the distributions shown in Table~\ref{table:free-flow-params}. The agents quickly separate due to relatively large time separation and minimum allowable separation values. The full scene evolution was created using Applied Intuition simulation tools and can be found on the lab video channel at \url{https://youtu.be/_HQSaUsixlM}.}}
\label{fig:free_flow_evolution}
\end{figure*}

\liam{Table~\ref{table:congested-params} and Table~\ref{table:free-flow-params} show the stochastic IDM parameters used to generate congested and free-flow traffic scenarios, respectively. Parameter selection was guided by the experimental values presented in  \cite{treiber2017intelligent, bhattacharyya2020online}. A demonstrative traffic scene consisting of 16 agents was created using the Applied Intuition simulation engine used by several companies for autonomy; the initial scene is displayed in \cref{fig:con00} and \cref{fig:ff00}. The governing rule-based model parameters were randomly selected for each agent according to $\sim \mathcal{N} (\mu,\sigma)$ for every parameter in Table~\ref{table:congested-params} and Table~\ref{table:free-flow-params}. Both scenes were then allowed to evolve for 60 seconds according to the outputs of seeded stochastic IDM models.} 

\begin{table}[b]
\begin{center}
 \begin{tabular}{llll} 
 \toprule
 \multicolumn{1}{l}{\textbf{Parameter}} & 
  \multicolumn{1}{l}{\textbf{Symbol}} &
  \multicolumn{1}{c}{$\mathbf{\mu}$} & 
  \multicolumn{1}{c}{$\mathbf{\sigma}$}
 \\ 
 \midrule
 Free flow speed & $v_{\mathrm{des}}$ & 16.0 & 1.5\\ 
 Minimum allowable separation & $d_{\mathrm{min}}$ & 3.0 & 0.5\\
 Minimum time separation & $\tau$ & 1.0 & 0.2\\
 Acceleration limit & $a_{\mathrm{max}}$ & 1.5 & 0.3\\
 Deceleration limit & $ b_{\mathrm{pref}}$ & 9.0 & 0.5\\
 IDM stochasticity parameter & $ \sigma_{\mathrm{IDM}}$ & 0.5 & 0.1\\
 \bottomrule
\end{tabular}
\end{center}
\caption{\liam{Sample driving parameters governing a congested traffic scenario.}}
\label{table:congested-params}
\end{table}

\begin{table}[b]
\begin{center}
 \begin{tabular}{llll} 
 \toprule
 \multicolumn{1}{l}{\textbf{Parameter}} & 
  \multicolumn{1}{l}{\textbf{Symbol}} &
  \multicolumn{1}{c}{$\mathbf{\mu}$} & 
  \multicolumn{1}{c}{$\mathbf{\sigma}$}
 \\ 
 \midrule
 Free flow speed & $v_{\mathrm{des}}$ & 29.0 & 2.5\\ 
 Minimum allowable separation & $d_{\mathrm{min}}$ & 5.0 & 1.0\\
 Minimum time separation & $\tau$ & 5.0 & 1.0\\
 Acceleration limit & $a_{\mathrm{max}}$ & 3.0 & 0.5\\
 Deceleration limit & $b_{\mathrm{pref}}$ & 9.0 & 0.5\\
 IDM stochasticity parameter & $\sigma_{\mathrm{IDM}}$ & 0.25 & 0.05\\
 \bottomrule
\end{tabular}
\end{center}
\caption{\liam{Sample driving parameters governing a free-flow traffic scenario.}}
\label{table:free-flow-params}
\end{table}

\liam{\Cref{fig:congested_evolution} displays the temporal evolution of the congested scene from an ego-centric perspective. The agents display a distinct clustering tendency that arises due to the relatively small time separation and minimum allowable separation values centered around the Table~\ref{table:congested-params} values; half of the agents visible in the first frame are still in-frame 45 seconds into the simulation. Such a scene could represent an urban driving scenario or congested traffic during a metropolitan rush hour.}

\liam{\Cref{fig:free_flow_evolution} displays the temporal evolution of the free-flow scene. Agents quickly gain separation to satisfy the relatively high time separation and minimum allowable separation values. Gaps between agents are noticeably longer than the gaps present in \cref{fig:congested_evolution}, and fewer agents remain visible in frames on average. This free-flow traffic scene could be leveraged to test an autonomous vehicle operating on a highway or in a low-traffic scenario.}

\liam{Kernel density estimates for aggregate vehicle speed and separation distance distributions are shown in \cref{fig:speed_kde} and \cref{fig:dist_kde}. These graphics succinctly demonstrate the powerful ability of rule-based driver models to generate diverse and interpretable macroscopic agent behavior via the judicious selection of a handful of parameters.}

\input{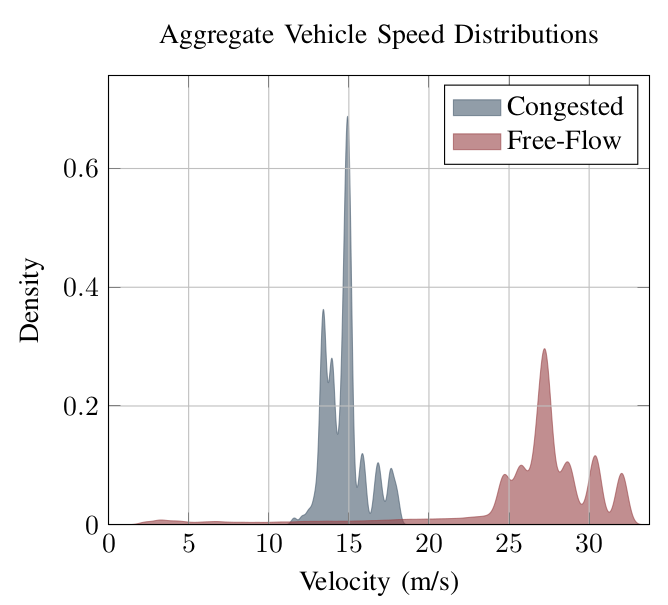}

\input{figs/kde/dist_kde}

\begin{figure*}[h]
    \begin{subfigure}[b]{0.48\textwidth}
       \includegraphics[width=\columnwidth]{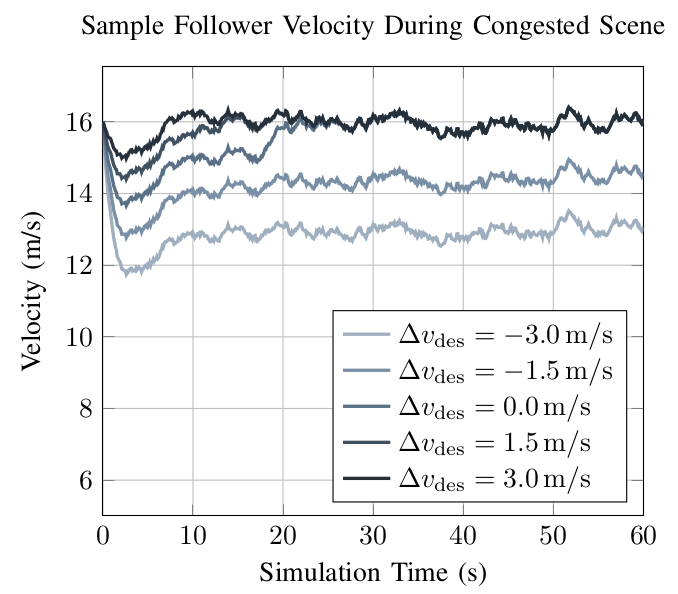}
        \caption{Congested scene.}
        \label{fig:follower_velocity_congested}
    \end{subfigure}
    \hfill
    \begin{subfigure}[b]{0.48\textwidth}
        \includegraphics[width=\columnwidth]{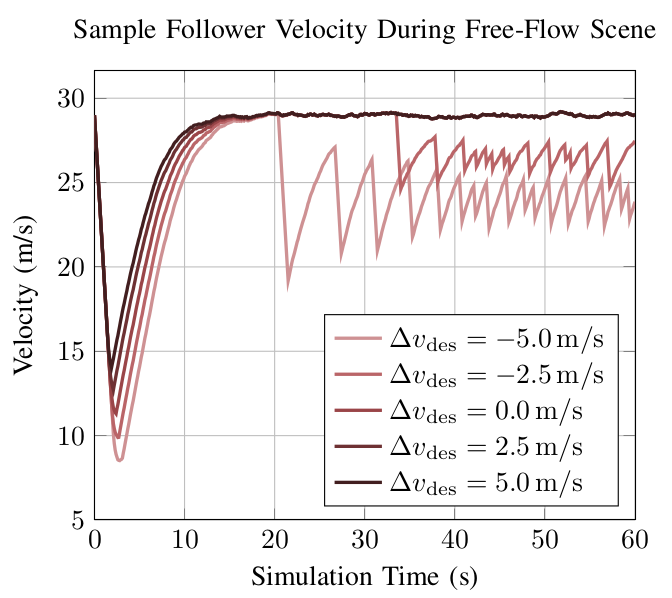}
        \caption{Free-flow scene.}
        \label{fig:follower_velocity_free_flow}
    \end{subfigure}
    \hfill
\caption{\liam{Follower velocities over a range of $\Delta v_{\mathrm{des}}$ values for both congested and free-flow scenarios. Note the oscillatory behavior that arises as the follower speeds up to achieve its desired free flow speed and then drops back to maintain its time separation objective.}}
\label{fig:follower_velocity}
\end{figure*}

\begin{figure*}[h]
    \begin{subfigure}[b]{0.48\textwidth}
        \includegraphics[width=\columnwidth]{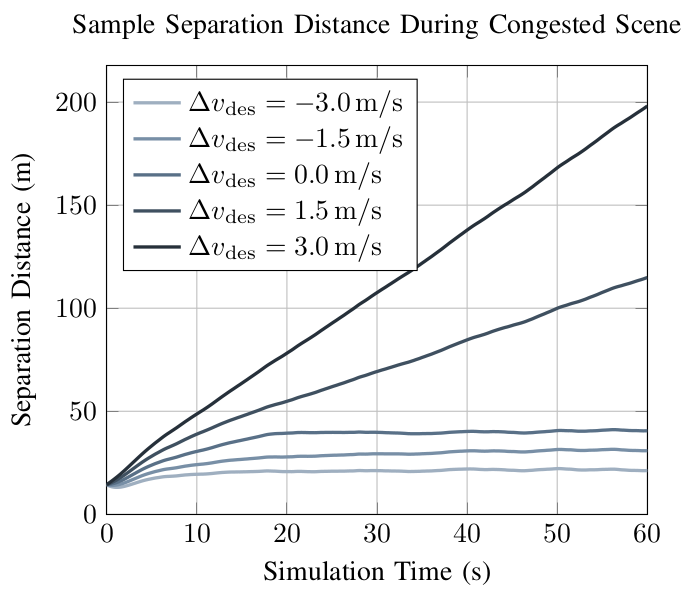}
        \caption{Congested scene.}
        \label{fig:sep_dist_congested}
    \end{subfigure}
    \hfill
    \begin{subfigure}[b]{0.48\textwidth}
        \includegraphics[width=\columnwidth]{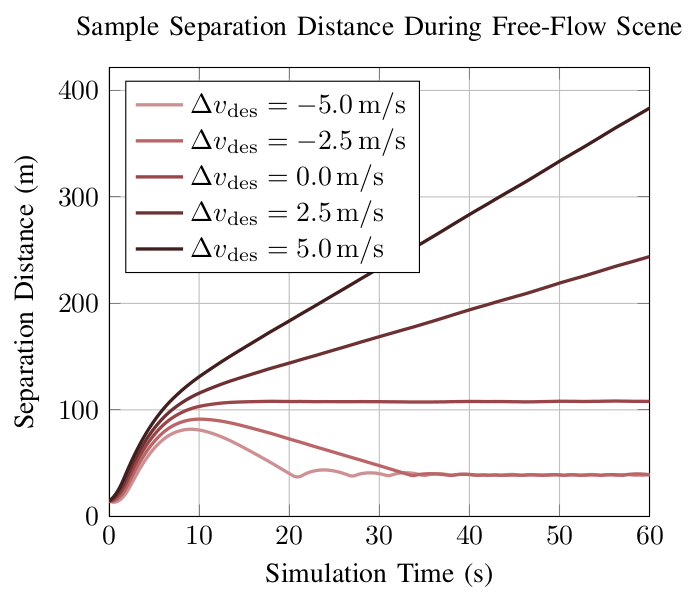}
        \caption{Free-flow scene.}
        \label{fig:sep_dist_free_flow}
    \end{subfigure}
\caption{\liam{Leader/follower separation distances over a range of $\Delta v_{\mathrm{des}}$ values for both congested and free-flow scenarios. Once again, note the oscillatory behavior that arises in the free-flow scene.}}
\label{fig:leader_follower_sep_dist}
\end{figure*}

\liam{A key determinant in scene evolution that deserves additional consideration is leader/follower behavior arising from differences in free flow speed. If the leader has free flow speed $v_{\mathrm{des},\ell}$ and the follower has free flow speed $v_{\mathrm{des},\mathit{f}}$, then we define the difference in free flow speed as $\Delta v_{\mathrm{des}} = v_{\mathrm{des},\ell} - v_{\mathrm{des},\mathit{f}}$. We explore the effect of $\Delta v_{\mathrm{des}}$ on vehicle behavior by isolating a leader/follower pair and setting their stochastic IDM parameter values to the mean values shown in Table~\ref{table:congested-params} and Table~\ref{table:free-flow-params}. The leader's $v_{\mathrm{des}}$ parameter is then swept over a range of values to induce a $\Delta v_{\mathrm{des}}$ with the follower. Follower velocities and leader/follower separation distances are shown in \cref{fig:follower_velocity} and \cref{fig:leader_follower_sep_dist}, respectively.}

\liam{The follower speed is effectively limited by the free flow speed of the leader. Recall that the follower has a desired free flow speed of $v_{\mathrm{des}} = \SI{16.0}{\meter/\second}$ in the congested scene and a desired free flow speed of $v_{\mathrm{des}} = \SI{29.0}{\meter/\second}$ in the free-flow scene. Negative $\Delta v_{\mathrm{des}}$ values force the follower to maintain a free flow speed below its desired value. This manifests itself in oscillatory follower behavior in the free-flow scenario seen in \cref{fig:follower_velocity_congested}, as the follower speeds up to achieve its desired free flow speed and then drops back to maintain its time separation objective. The time separation objective also produces the pronounced drop in follower velocity at the start of the scene, as the follower vehicle waits for the leader to establish the desired separation. This outcome is interpretable and expected due to the rule-based nature of the driving scene, but is not altogether representative of real-world driving behavior. Thus, simulation scenes should be allowed to reach steady-state behavior or should be initialized with adequate separation distance between leader/follower pairs to avoid a dramatic braking maneuver. The pronounced braking behavior is visible around 00:03 in the free-flow video shown in \url{https://youtu.be/_HQSaUsixlM}.}

\liam{Similar trends arise in the separation distance plots shown in \cref{fig:leader_follower_sep_dist}. Leader/follower separation distance is ultimately bound by the differences in free flow speed if the follower is seeded with a larger $v_{\mathrm{des}}$ value than the leader. A slight oscillation is detectable in the free-flow separation distance seen in \cref{fig:sep_dist_free_flow}, as the follower faces competing objectives due to its time separation and free-flow speed objectives.}

\liam{In future work, datasets for distinct driving scenarios could be economically generated using different parameter seeds for the stochastic IDM model. The methods presented in this work could be then further compared across traffic conditions to identify scenarios that highlight the strengths and weaknesses of a particular approach.}

\section{Conclusion}
Reliable models of human driving are essential for the safety validation of autonomous driving algorithms.
In this paper, we described a hybrid rule-based and data-driven method to model human driving from demonstrations.
Using the well-understood technique of particle filtering, we inferred distributions over the parameters governing underlying rule-based driver models.
While the rule-based models provide interpretable driving behavior, the data-driven parameter estimation provides fidelity to real-world driving demonstrations.

We conducted experiments using driving demonstrations from three real-world driving datasets: NGSIM~\cite{colyar2007us}, HighD~\cite{highDdataset}, and Interaction~\cite{interactiondataset} on the driving tasks of highway driving and merging.
We baselined the driver models obtained using our method against both rule-based and black-box driver models and showed that our model is better able to capture the real-world driving behavior in rollout experiments.
While rollout experiments establish the closeness of generated trajectories to demonstrations, we also assessed the trajectory generation capability of our models by using them to generate novel trajectories.
We conducted a ``driving Turing test'' by showing videos of generated trajectories to human volunteers. 
The test confirmed that our models were able to generate realistic traffic behavior.

There are interesting directions for future work. While we assumed fixed parameters of the underlying rule-based models, future work will investigate the impact of changing parameters to account for nonstationarity in human driving behavior. 
Going forward, we also plan to deploy our method on more diverse applications of human behavior modeling. One such avenue is using our method in combination with rule-based models that rely on graph structures that can model more complex interactions such as probabilistic pedestrian motion modeling~\cite{jain2020discrete} and crowd modeling~\cite{li2020pedestrian}.
Finally, this paper did not tackle the question of when model-based and model-free approaches should be used to model driving behavior.
Future work could build a suite of driving scenarios and assess which kind of model is better suited to a given situation.

\section*{Acknowledgments}
Toyota Research Institute (TRI) provided funds to assist the authors with their research, but this article solely reflects the opinions and conclusions of its authors and not TRI or any other Toyota entity.

% \section*{REFERENCES}
% \renewcommand{\section}[2]{}%
% \bibliography{ref.bib}

% \clearpage
% \newpage

% \bibliographystyle{IEEEtran}
% \bibliography{ref}

\renewcommand*{\bibfont}{\footnotesize}
\printbibliography

% \addtolength{\textheight}{-12cm}   
% This command serves to balance the column lengths on the last page of the document manually. It shortens the textheight of the last page by a suitable amount. This command does not take effect until the next page so it should come on the page before the last. Make sure that you do not shorten the textheight too much.

\end{document}